# A Survey on Artificial Intelligence Assurance

Feras A. Batarseh; Bradley Department of Electrical and Computer Engineering, Virginia Polytechnic Institute and State University (Virginia Tech), Arlington, VA 22203
(*Corresponding author) batarseh@vt.edu

Laura Freeman; Department of Statistics, Virginia Polytechnic Institute and State University (Virginia Tech), Arlington, VA 22203 laura.freeman@vt.edu

Chih-Hao Huang; College of Science, George Mason University, Fairfax 22030 chuang21@gmu.edu

**Abstract** – Artificial Intelligence (AI) algorithms are increasingly providing decision making and operational support across multiple domains. AI includes a wide library of algorithms for different problems. One important notion for the adoption of AI algorithms into operational decision process is the concept of assurance. The literature on assurance, unfortunately, conceals its outcomes within a tangled landscape of conflicting approaches, driven by contradicting motivations, assumptions, and intuitions. Accordingly, albeit a rising and novel area, this manuscript provides a systematic review of research works that are relevant to AI assurance, between years 1985 – 2021, and aims to provide a structured alterative to the landscape. A new AI assurance definition is adopted and presented and assurance methods are contrasted and tabulated. Additionally, a ten-metric scoring system is developed and introduced to evaluate and compare existing methods. Lastly, in this manuscript, we provide foundational insights, discussions, future directions, a roadmap, and applicable recommendations for the development and deployment of AI assurance.

## 1. Introduction and survey structure

The recent rise of big data gave birth to a new promise for AI based in statistical learning, and at this time, contrary to previous AI winters, it seems that statistical learning enabled AI has survived the hype, in that it has been able to surpass human-level performance in certain domains. Similar to any other engineering deployment, building AI systems requires evaluation, which may be called assurance, validation, verification or another name. We address this terminology debate in the next section.

Defining the scope of AI assurance is worth studying, AI is currently deployed at multiple domains, it is forecasting revenue, guiding robots in the battlefield, driving cars, recommending policies to government officials, predicting pregnancies, and classifying customers. AI has multiple subareas such as machine learning, computer vision, knowledge-based systems, and many more – therefore, we pose the question: is it possible to provide a generic assurance solution across all subareas and domains? This review sheds light on existing works in AI assurance, provides a comprehensive overview of the *state-of-the-science*, and discusses patterns in AI assurance publishing.. This section sets that stage for the manuscript by presenting the motivation, clear definitions and distinctions, as well as the inclusion/exclusion criteria of reviewed articles.

## 1.1 Relevant terminology and definitions

All AI systems require assurance; it is important to distinguish between different terms that might have been used interchangeably in literature. We acknowledge the following relevant terms: (1) validation, (2) verification, (3) testing, and (4) assurance. This paper is concerned with all of the mentioned terms. The following definitions are adopted in our manuscript, for the purposes of clarity and to avoid ambiguity in upcoming theoretical discussions:

**Verification**: "The process of evaluating a system or component to determine whether the products of a given development phase satisfy the conditions imposed at the start of that phase". **Validation**: "The process of evaluating a system or component during or at the end of the development process to determine whether it satisfies specified requirements" (Gonzalez and Barr, 2020). Another definition for V&V is from the Department of Defense, as they applied testing practices to simulation systems, it states the following: Verification is the "process of determining that a model implementation accurately represents the developer's conceptual descriptions and specifications", and Validation is the process of "determining the degree to which a model is an accurate representation" (DoD, 1995).

**Testing**: according to the American Software testing Qualification Board, testing is "the process consisting of all lifecycle activities, both static and dynamic, concerned with planning, preparation and evaluation of software products and related work products to determine that they satisfy specified requirements, to demonstrate that they are fit for purpose and to detect defects". Based on that (and other reviewed definitions), testing includes both validation and verification.

**Assurance**: this term has been rarely applied to conventional software engineering; rather, it is used in the context of AI and learning algorithms. In this manuscript, based on prior definitions and recent AI challenges, we propose the following definition for AI assurance:

*A process that is applied at all stages of the AI engineering lifecycle ensuring that any intelligent system is producing outcomes that are valid, verified, data-driven, trustworthy and explainable to a layman, ethical in the context of its deployment, unbiased in its learning, and fair to its users.*

Our definition is by design generic and therefore applicable to all AI domains and subareas. Additionally, based on our review of a wide variety of existing definitions of assurance, it is evident that the two main AI components of interest are *the data* and *the algorithm*; accordingly, those are the two main pillars of our definition. Additionally, we highlight that the outcomes the AI enable system (intelligent system) are evaluated at the system level, where the decision or action is being taken.

The remaining of this paper is focused on a review of existing AI assurance methods, and it is structured as follows: the next section presents the inclusion/exclusion criteria, section 2 provides a historical perspective as well as the entire assurance landscape, section 3 includes an exhaustive list of papers relevant to AI assurance (as well as the scoring system), section 4 presents overall insights and discussions of the survey, and lastly, section 5 presents conclusions.1.2 Description of included articles

Articles that are included in this paper were found using the following search terms: assurance, validation, verification, and testing. Additionally, as it is well known, AI has many subareas, in this paper, the following subareas were included in the search: machine learning, data science, deep learning, reinforcement learning, genetic algorithms, agent-based systems, computer vision, natural language processing, and knowledge-based systems (expert systems). We looked for papers in conference proceedings, journals, books and book chapters, dissertations, as well as industry white papers. The search yielded results from year 1985 to year 2021. Besides university libraries, multiple online repositories were searched (the most commonplace AI peer-reviewed venues). Additionally, areas of research such as data bias, data incompleteness, Fair AI, Explainable AI (XAI), and Ethical AI were used to widen the net of search. The next section presents an executive summary of the history of AI assurance.

## 2. AI assurance landscape

The history and current state of AI assurance is certainly a debatable matter. In this section, multiple methods are discussed, critiqued, and aggregated by AI subarea. The goal is to illuminate the need for an organized system for evaluating and presenting assurance methods; which is presented in next sections of this manuscript.

### 2.1 A historical perspective (analysis of the state-of-the-science)

As a starting point for AI assurance and testing, there is nowhere more suitable to begin than the Turing test (Turing 1950). In his famous manuscript: Computing Machinery and Intelligence, he introduced the imitation game, which was then popularized as the Turing test. Turing states: "The object of the game for the interrogator is to determine which of the other two is the man and which is the woman". Based on a series of questions, the intelligent agent "learns" how to make such a distinction. If we consider the different types of intelligence, it becomes evident that different paradigms have different expectations. A genetic algorithm aims to optimize, while a classification algorithm aims to classify (choose between yes and no for instance). As Turing stated in his paper: "We are of course supposing for the present that the questions are of the kind to which an answer: Yes or No is appropriate, rather than questions such as: What do you think of Picasso?" Comparing predictions (or classifications) to actual outputs is one way of evaluating that the results of an algorithm match what the real world created.

There were a dominating number of validation and verification methods in the seventies, eighties, and nineties for two forms of intelligence, knowledge-based systems (i.e., expert systems) and simulation systems (majorly for defense and military applications). One of the first times where AI turned towards data-driven methods was apparent in 1996 at the Third International Math and Science Study (TIMSS), which , focused on quality assurance in data collection (Martin and Mullis, 1996). Data from Forty-five countries were included in the analysis. In a very deliberate process, the data collectors were faced with challenges relevant to the internationalization of data. For example, data from Indonesia had errors in translation; data collection processes were different in Korea, Germany, and Kuwait than the standard process due to funding and timing issues. Such real-world issues in data collection certainly pose a challenge to the assurance of statistical learning AI that require addressing.

In the 1990s, AI testing and assurance were majorly inspired by the big research archive of testing of software (i.e., within software engineering) (Batarseh et al., 2020). However, a slim amount of literature explored algorithms such as genetic algorithms (Jones et al., 1997), reinforcement learning (Hailu and Sommer, 1997), and neural networks (Paladini, 1999). It was not until the 2000s that there was a serious surge in data-driven assurance and the testing of AI methods.

In the early 2000s, mostly manual methods of assurance were developed, for example, CommonKADS was a popular and commonplace method that was used to incrementally develop and test an intelligent system. Other domain-specific works were published in areas such as healthcare (Berndt et al., 2001), or algorithms-specific assurance such as Crisp Clustering for k-means clustering (Halkidi et al., 2001).

It was not until the 2010s that a spike in AI assurance for *big* data occurred. Validation of data analytics and other new areas, such as XAI and Trustworthy AI have dominated the AI assurance field in recent years. Figure 1 illustrates that areas including XAI, computer vision, deep learning, and reinforcement learning have had a recent spike in assurance methods; and the trend is expected to be increasingly on the rise (as shown in Figure 2). The figure also illustrates that knowledge-based systems were the focus until the early nineties, and shows a shift towards the statistical learning based subareas in the 2010s. A version of the dashboard is available in a public repository (with instructions on how to run it): https://github.com/ferasbatarseh/AI-Assurance-Review

The p-values for the trend lines presented in Figure 2 are as follows: Data Science (DS): 0.87, Genetic Algorithms (GA): 0.50, Reinforcement Learning (RL): 0.15, Knowledge-Based Systems (KBS): 0.97, Computer Vision (CV): 0.22, Natural Language Processing (NLP): 0.17, Generic AI: 0.95, Agent-Based Systems (ABS): 0.33, Machine Learning (ML): 0.72, Deep Learning (DL): 0.37, and XAI: 0.44.

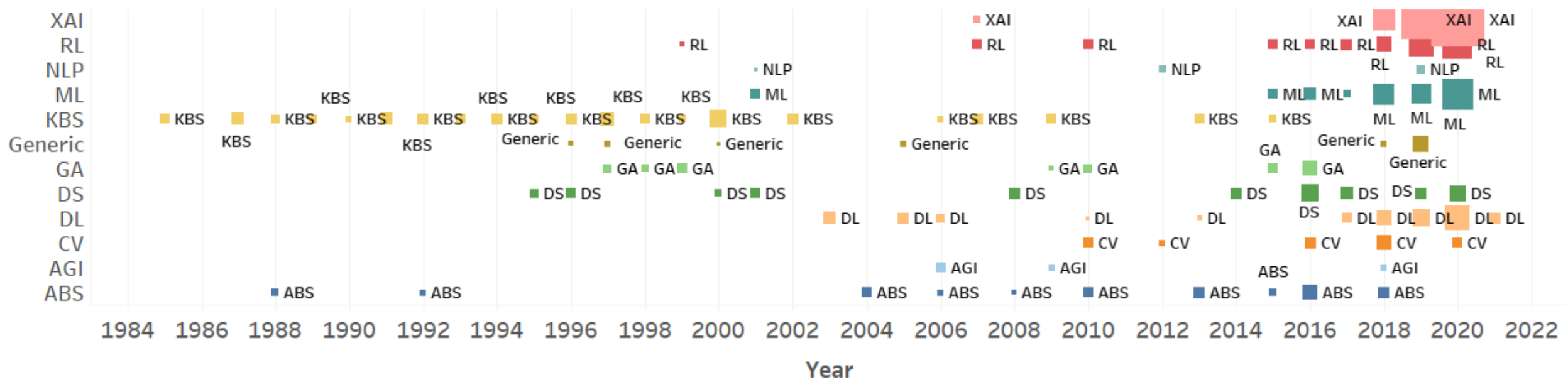

Figure 1: A history of AI assurance by year and subarea

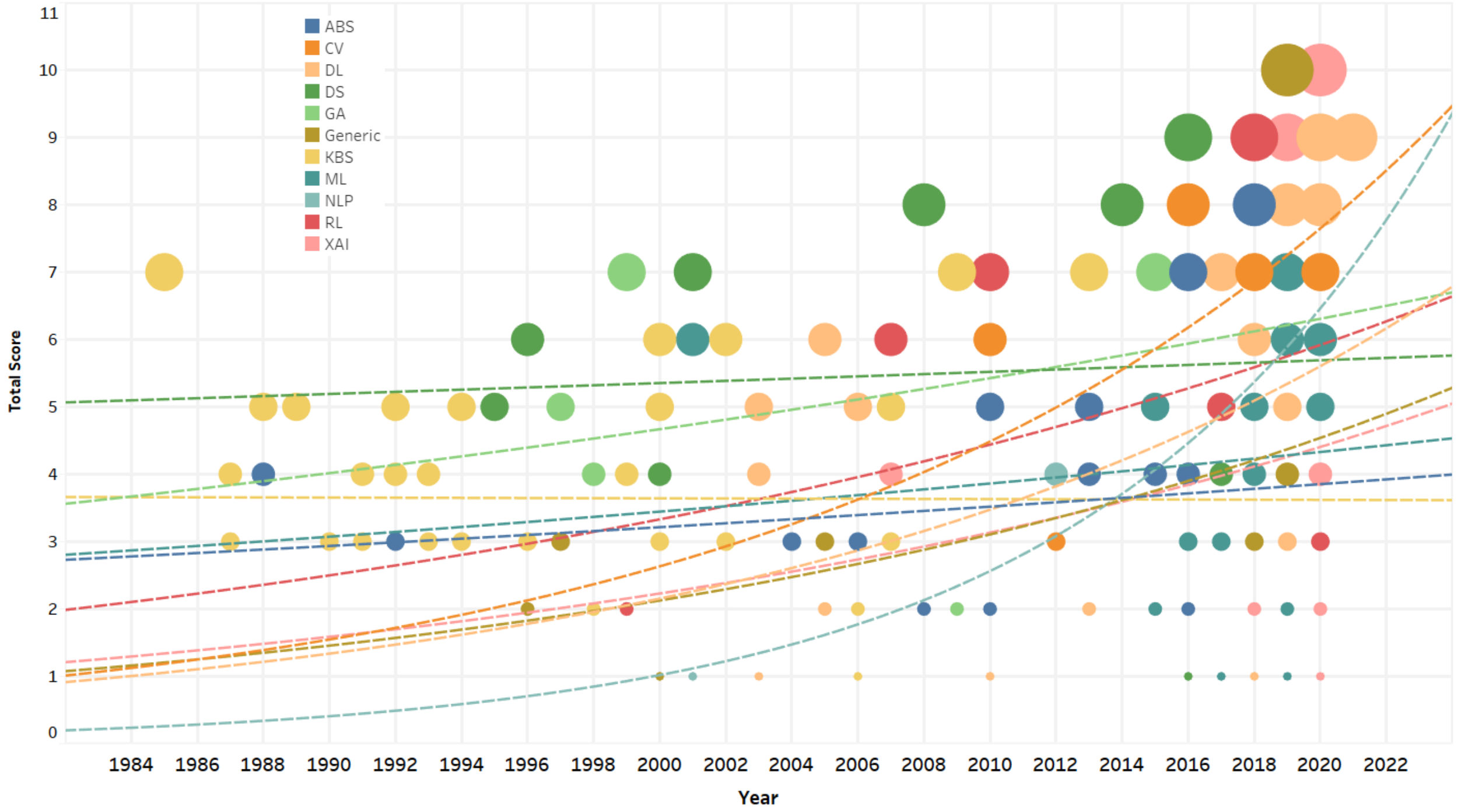

Figure 2: Past and future (using trend lines) of AI assurance (by AI subarea)

It is undeniable that there is a rise in the research of AI, and especially in the area of assurance. The next section (2.2) provides further details on the state-of-the-art, and section 3 presents an exhaustive review of all AI assurance methods found under the predefined search criteria.

## 2.2 The state of AI assurance

This section introduces some milestone methods and discussion in AI assurance. Many of the discussed works rely on standard software validation and verification methods. Such methods are inadequate for AI systems, because they have a dimension of intelligence, learning, and re-learning, as well as adaptability to certain contexts. Therefore, errors in AI system "may manifest themselves because of autonomous changes" (Taylor, 2006), and among other scenarios would require extensive assurance. For instance, in expert systems, the inference engine component creates rules and new logic based on forward and backward propagation (Batarseh & Gonzalez, 2013). Such processes require extensive assurance of the process as well as the outcome rules. Alternatively, for other AI areas such as neural networks, while propagation is used, taxonomic evaluations and adversarial targeting are more critical to their assurance (Massoli et al., 2021). For other subareas such as machine learning, the structure of data, data collection decisions, and other data-relevant properties need step-wise assurance to evaluate the resulted predictions and forecasts. For instance, several types of bias can occur in any phase of the data science lifecycle or while extracting outcomes. Bias can begin during data collection, data wrangling, modeling, or any other phase. Biases and variances which arise in the data are independent of the sample size or statistical significance, and they can directly affect the *context* or the results or the model. Other issues such as incompleteness, data skewness, or lack of structure have a negative influence on the quality of outcomes of any AI model and require data assurance (Kulkarni et al., 2020).

While the historic majority of methods for knowledge-based systems and expert systems (as well as neural networks) aimed at finding generic solutions for their assurance (Tsai et al, 1999), (Batarseh & Gonzalez, 2015), and (Onoyama & Tsuruta, 2000), other "more recent" methods were focused on one AI subarea and one domain. For instance, in Mason et al. (2017), assurance was applied to reinforcement learning methods for safety-critical systems. Precentzas et al. (2019) presented an assurance method for machine learning as its applied to stroke predictions, similar to Pawar's et al's (2020) XAI for healthcare framework. Pepe et al. (2009), and Chittajallu et al.'s (2019) developed a method for surgery video detection methods. Moreover, domains such as law and society would generally benefit from AI subareas such as natural language processing for analyzing legal contracts (Magazzeni, 2017), but also require assurance.

Another major aspect (for most domains) that was evident in the papers reviewed was the need for explainability (i.e. XAI) of the learning algorithm, defined as: *to identify how the outcomes were arrived at* (transforming the black-box to a white-box) (Schlegel et al., 2019). Few papers without substantial formal methods were found for Fair AI, Safe AI (Everitt, 2018), Transparent AI (Behnoush & Nasraoui, 2018), or Trustworthy AI (Aitken et al., 2016); but XAI (Hagras, 2018) has been central (as the previous figures in this paper also suggest). For instance, in Lee et al. (2019), layer-wise relevance propagation was introduced to obtain the effects of every neural layer and each neuron on the outcome of the algorithm. Those observations are then presented for better understanding of the model and its inner workings. Additionally, Arrieta et al. (2019) presented a model for XAI that is tailored for road traffic forecasting, and Guo (2020) presented the same, albeit for 5G and wireless networks (Spada & Vincentini, 2019). Similarly, Kuppa and Le-Khac (2020) presented a method focused on Cyber Security using gradient maps and bots. Go & Lee (2018) presented an AI assurance method for trustworthiness of security systems. Lastly, Guo (2020) developed a framework for 6G testing using deep neural networks.

Multi-agent AI is another aspect that requires a specific kind of assurance, by validating every agent, and verifying the integration of agents (Nourani et al., 2016). The challenges of AI algorithms and their assurance is evident and consistent across many of the manuscripts, such as in Janssen and Kuk's (2016) study of the limitations of AI for government, on the other hand, Batarseh et al. (2017) presented multiple methods for applying data science at government (with assurance using knowledge-based systems). Assurance is especially difficult when it comes to being performed in real time; timeliness in critical systems, and other defense-relevant environments is very important (Jorge et al., 2018), (Bruno et al., 2017), and (Laat, 2017). Other less "time-constrained" activities such as decisions at organizations (Ruan, 2017) and time series decision support systems could utilize slower methods such as genetic algorithms (Thomas & Sycara, 1999), but they require a different take on assurance. The authors suggested that "by no means we have a definitive answer, what we do here is intended to be suggestive" (Thomas & Sycara, 1999) when addressing the validation part of their work. A recent publication by Raji et al. (2020) shows a study from the Google team claiming that they are "aiming to close the accountability gap of AI" using an internal audit system (at Google). IBM research also proposed few solutions to manage the bias of AI services (Srivastava & Rossi, 2019) (Varshney, 2020). As expected, the relevance and success of assurance methods varied, and so we developed a scoring system to evaluate existing methods. We were able to identify 200+ relevant manuscripts with methods. The next section presents the exhaustive list of the works presented in this section in addition to multiple others with our derived scores.

## 3. The review and scoring of methods

The scoring of each AI assurance method/paper was based on the sum of the score of ten metrics. The objective of the metrics is to provide readers with a meaningful strategy for sorting through the vast literature on AI assurance. The scoring metric is based on the authors' review of what makes a useful reference paper for AI assurance. Each elemental metric is allocated one point, and each method is either given that point or not (0 or 1), as follows:

I. Specificity to AI: some assurance methods are generically tailored to many systems, others are deployable only to *intelligent* systems; one point was assigned to methods that focused (i.e. specific) on the inner workings of AI systems.
II. The existence of a formal method: this metric indicates whether the manuscript under review presented a formal (quantitative and qualitative) description of their method (1 point) or not (0 points).
III. Declared successful results: in experimental work of a method under review, some authors declared success and presented success rates, if that is present, we gave that method a point.
IV. Datasets provided: whether the method has a big dataset associated with it for testing (1) or not (0). This is an important factor for reproducibility and research evaluation purposes.

V. AI system size: methods were applied to a small AI system, other were applied to bigger systems for instance, we gave a point to methods that could be applied to big real-world systems rather than ones with theoretical deployments.

VI. Declared success: whether the authors declared success of their method in reaching an *assured* AI system (1) or not (0).

VII. Mentioned limitations: whether there are obvious method limitations (0) or not (1).

VIII. Generalized to other AI deployments: some methods are broad and are able to be generalized for multiple AI systems (1), others are "narrow" (0) and more specific to one application or one system.

IX. A real-world application: if the method presented is applied to a real-world application, it is granted one point.

X. Contrasted with other methods: if the method reviewed is compared, contrasted, or measured against other methods, or if it proves its superiority over other methods, then it is granted a point.

Table 1 presents the methods reviewed, along with their first author's last name, publishing venue, AI subarea, as well as the score (sum of ten metrics).

Other aspects such as domain of application were missing from many papers and inconsistent, therefore, we didn't include them in the table. Additionally, we considered *citations per paper*. However, the data on citations (for a 250+ papers study) were incomplete and difficult to find in many cases. For many of the papers, we did not have information on how many times they were cited, because many publishers failed to index their papers across consistent venues (e.g., Scopus, MedLine, Web of Science, and others). Additionally, the issue of *self-citation* is in some cases considered in scoring but in other cases is not. Due to these citation inconsistencies (which are believed to be a challenge that reaches all areas of science), we deemed that using citations would provide more questions than answers than our subject matter expert based metrics.

Appendix 1 presents a list of all reviewed manuscripts and their detailed scores (for the ten metrics) by ranking category. The papers, data, dashboard, and lists are on a public GitHub repository: https://github.com/ferasbatarseh/AI-Assurance-Review

**Table 1: Reviewed methods and their scores**

| Year | First Author's Last Name and Citation | Publishing Venue | AI Subarea | Total Score |
|---|---|---|---|---|
| **2020** | D'Alterio (D'Alterio et al., 2020) | FUZZ-IEEE | XAI | 10 |
| **2019** | Tao (C. Tao et al., 2019) | IEEE Access | Generic | 10 |
| **2020** | Anderson (A. Anderson et al., 2020) | ACM TIIS | RL | 9 |
| **2020** | Birkenbihl (Birkenbihl, 2020) | EPMA | ML | 9 |
| **2020** | Checco (Checco et al., 2020) | JAIR | DS | 9 |
| **2020** | Chen (H.-Y. Chen & Lee, 2020) | IEEE Access | XAI | 9 |
| **2020** | Cluzeau (Cluzeau et al., 2020) | EASA | DL | 9 |
| **2019** | Kaur (Kaur et al., 2019) | WAINA | XAI | 9 |
| **2020** | Kulkarni (Kulkarni et al., 2020) | Academic Press | DS | 9 |
| **2020** | Kuppa (Kuppa & Le-Khac, 2020) | IEEE IJCNN | XAI | 9 |
| **2020** | Kuzlu (Kuzlu et al., 2020) | IEEE Access | XAI | 9 |
| **2021** | Massoli (Massoli et al., 2021) | CVIU | DL | 9 |
| **2020** | Spinner (Spinner et al., 2019) | IEEE TVCG | XAI | 9 |
| **2016** | Veeramachaneni (Veeramachaneni et al., 2016) | IEEE HPSC | DS | 9 |
| **2018** | Wei (Wei et al., 2018) | AS | RL | 9 |
| **2020** | Winkel (Winkel, 2020) | EJR | RL | 9 |
| **2014** | Ali (Ali & Schmid, 2014) | GISci | DS | 8 |
| **2018** | Alves (Alves et al., 2018) | NASA ARIAS | ABS | 8 |
| **2019** | Batarseh (Batarseh & Kulkarni, 2019) | EDML | DS | 8 |
| **2016** | Gao (Gao et al., 2016) | SEKE | DS | 8 |
| **2020** | Gardiner (Gardiner et al., 2020) | Nature Sci Rep | ML | 8 |
| **2016** | Gulshan (Gulshan et al., 2016) | JAMA | CV | 8 |
| **2020** | Guo (Guo, 2020a) | IEEE ICCVW | XAI | 8 |
| **2020** | Han (Han et al., 2020) | IET JoE | XAI | 8 |
| **2016** | Heaney (Heaney et al., 2016) | OD | GA | 8 |
| **2019** | Huber (Huber, 2019) | KI AAI | RL | 8 |
| **2019** | Keneni (Keneni et al., 2019) | IEEE Access | XAI | 8 |
| **2020** | Kohlbrenner (Kohlbrenner et al., 2020) | IEEE IJCNN | XAI | 8 |
| **2019** | Maloca (Maloca et al., 2019) | PLoS ONE | DL | 8 |
| **2020** | Malolan (Malolan et al., 2020) | IEEE ICICT | XAI | 8 |
| **2020** | Payrovnaziri (Payrovnaziri et al., 2020) | JAMIA | ML | 8 |
| **2008** | Peppler (Peppler et al., 2008) | OASJ | DS | 8 |
| **2020** | Sequeira (Sequeira & Gervasio, 2020) | SciDir AI | RL | 8 |
| **2020** | Sivamani (Sivamani et al., 2020) | IEEE LCS | DL | 8 |
| **2020** | Tan (Tan et al., 2020) | IEEE IJCNN | XAI | 8 |
| **2020** | Tao (J. Tao et al., 2020) | IEEE CoG | XAI | 8 |
| **2020** | Welch (Welch et al., 2020) | PhysMedBiol | DL | 8 |
| **2020** | Xiao (Xiao et al., 2020) | IS | DL | 8 |
| **2016** | Aitken (Aitken, 2016) | UC | ABS | 7 |
| **2019** | Barredo-Arrieta (Barredo-Arrieta et al., 2019) | IEEE ITSC | XAI | 7 |
| **2013** | Batarseh (Batarseh & Gonzalez, 2013) | IEEE TSMCS | KBS | 7 |
| **2001** | Berndt (Berndt et al., 2001) | COMP | DS | 7 |
| **2010** | Bone (Bone & Dragićević, 2010) | CEUS | RL | 7 |
| **2016** | Celis (Celis et al., 2016) | PrePrint | ML | 7 |
| **2019** | Chittajallu (Chittajallu et al., 2019) | IEEE ISBI | XAI | 7 |
| **2018** | Elsayed (Elsayed et al., 2018) | NIPS | CV | 7 |
| **2019** | Ferreyra (Ferreyra et al., 2019) | FUZZ-IEEE | XAI | 7 |
| **2006** | Forster (Forster, 2006) | Uni of South Africa | AGI | 7 |
| **1985** | Ginsberg (Ginsberg & Weiss, 2001) | IJCAI | KBS | 7 |
| **2018** | Go (Go & Lee, 2018) | ACM CCS | DL | 7 |
| **2020** | Halliwell (Halliwell & Lecue, 2020) | PrePrint | DL | 7 |
| **2015** | He (C. He et al., 2015) | MPE | GA | 7 |
| **2020** | Heuer (Heuer & Breiter, 2020) | ACM UMAP | ML | 7 |
| **2016** | Jiang (Jiang & Li, 2016) | PMLR | RL | 7 |
| **2020** | Kaur (Kaur et al., 2020) | AINA | XAI | 7 |
| **2016** | Kianifar (Kianifar, 2016) | SC | GA | 7 |
| **2019** | Lee (J. ha Lee et al., 2019) | IEEE ICTC | XAI | 7 |
| **2017** | Liang (Liang et al., 2017) | MILCOM | DS | 7 |
| **2020** | Mackowiak (Mackowiak et al., 2020) | PrePrint | CV | 7 |
| **2018** | Mason (Mason et al., 2018) | AHIM | RL | 7 |
| **2018** | Murray (B. Murray et al., 2018) | FUZZ-IEEE | XAI | 7 |
| **2019** | Naqa (El Naqa et al., 2019) | MedPhys | ML | 7 |
| **2019** | Prentzas (Prentzas et al., 2019) | IEEE BIBE | XAI | 7 |
| **2018** | Pynadath (Pynadath, 2018) | Springer HCIS | ML | 7 |
| **2020** | Ragot (Ragot et al., 2020) | CHI | ML | 7 |
| **2020** | Rotman (Rotman et al., 2020) | PrePrint | RL | 7 |
| **2015** | Rovcanin (Rovcanin et al., 2015) | WN | RL | 7 |
| **2020** | Sarathy (Sarathy et al., 2020) | IEEE SISY | XAI | 7 |
| **2018** | Stock (Stock & Cisse, 2018) | ECCV | CV | 7 |
| **2009** | Tadj (Tadj, 2005) | SCI | KBS | 7 |
| **1999** | Thomas (Thomas & Sycara, 1999) | AAAI | GA | 7 |
| **2020** | Uslu (Uslu et al., 2020a) | AINA | XAI | 7 |
| **2018** | Xu (Xu et al., 2018) | PrePrint | DL | 7 |
| **2019** | Bellamy (Bellamy et al., 2019) | IBM JRD | XAI | 6 |
| **2019** | Beyret (Beyret et al., 2019) | IEEE IROS | RL | 6 |
| **2018** | Cao (Cao et al., 2019) | JAIHC | ML | 6 |
| **2020** | Cruz (Cruz et al., 2020) | PrePrint | RL | 6 |
| **2001** | Halkidi (Halkidi et al., 2001) | JIIS | ML | 6 |
| **2020** | He (Y. He et al., 2020) | PrePrint | RL | 6 |
| **2020** | Islam (Islam et al., 2019) | IEEE TFS | XAI | 6 |
| **2005** | Liu (F. Liu & Yang, 2005) | AI2005 | DL | 6 |
| **2019** | Madumal (Madumal et al., 2019) | PrePrint | RL | 6 |
| **1996** | Martin (Martin et al., 1996) | ERIC | DS | 6 |
| **2007** | Martín-Guerrero (Martín-Guerrero et al., 2007) | AJCAI | RL | 6 |
| **2000** | Mosqueira-Rey (Mosqueira-Rey & Moret-Bonillo, 2000) | ESA | KBS | 6 |
| **2020** | Mynuddin (Mynuddin & Gao, 2020) | IETITS | RL | 6 |
| **2020** | Puiutta (Puiutta & Veith, 2020) | CD-MAKE | RL | 6 |
| **2018** | Ruan (Ruan et al., 2018) | IJCAI | DL | 6 |
| **2019** | Schlegel (Schlegel et al., 2019) | IEEE ICCVW | XAI | 6 |
| **2020** | Toreini (Toreini et al., 2020) | ACM FAT | ML | 6 |
| **2020** | Toreini (Toreini et al., 2020) | PrePrint | ML | 6 |
| **2019** | Vabalas (Vabalas et al., 2019) | PLoS ONE | ML | 6 |
| **2010** | Winkler (Winkler & Rinner, 2010) | IEEE SUTC | CV | 6 |
| **2002** | Wu (Wu & Lee, 2002) | IJHCS | KBS | 6 |
| **2019** | Zhu (H. Zhu et al., 2019) | ACM PLDI | RL | 6 |
| **1992** | Andert (Andert, 1992) | IJM | KBS | 5 |
| **2018** | Antunes (Antunes et al., 2018) | IEEE DSN-W | ML | 5 |
| **1989** | Becker (Becker et al., 1989) | NASA | KBS | 5 |
| **2019** | Chen (T. Chen et al., 2019) | CS | RL | 5 |
| **2019** | Cruz (Cruz et al., 2019) | AI 2019 AAI | RL | 5 |
| **2020** | Diallo (Diallo et al., 2020) | IEEE ACSOS-C | XAI | 5 |
| **2010** | Dong (Dong et al., 2010) | IEEE ICWIIAT | GA | 5 |
| **2019** | Dupuis (Dupuis & Verheij, 2019) | UoG | XAI | 5 |
| **2015** | Goodfellow (Goodfellow et al., 2015) | PrePrint | ML | 5 |
| **2020** | Guo (Guo, 2020b) | IEEE CM | XAI | 5 |
| **2020** | Haverinen (Haverinen, 2020) | Uni of Jyväskylä | XAI | 5 |
| **1997** | Jones (Jones et al., 1997) | JMB | GA | 5 |
| **2019** | Joo (Joo & Kim, 2019) | IEEE CoG | RL | 5 |
| **2020** | Katell (Katell et al., 2020) | ACM FAT | XAI | 5 |
| **2007** | Knauf (Rainer Knauf et al., 2007) | IEEE TSMC | KBS | 5 |
| **1995** | Lockwood (Lockwood & Chen, 1995) | AES | KBS | 5 |
| **2000** | Marcos (Marcos et al., 2000) | IEE Proc | KBS | 5 |
| **2017** | Mason (Mason et al., 2017b) | WhiteRose | RL | 5 |
| **1988** | Morell (Morell, 1988) | IEA/AIE | KBS | 5 |
| **2020** | Murray (B. J. Murray et al., 2020) | IEEE TETCI | XAI | 5 |
| **2010** | Niazi (Niazi et al., 2010) | SpringSim | ABS | 5 |
| **2000** | Onoyama (Onoyama & Tsuruta, 2000) | JETAI | KBS | 5 |
| **2019** | Ren (Ren et al., 2019) | PrePrint | DL | 5 |
| **2013** | Sargent (Robert G. Sargent, 2013) | JoS | ABS | 5 |
| **2003** | Schumann (Schumann et al., 2003) | EANN | DL | 5 |
| **1995** | Singer (Singer et al., 1995) | POQ | DS | 5 |
| **2019** | Srivastava (Srivastava & Rossi, 2019) | AAAI AIES | NLP | 5 |
| **2006** | Taylor (Brian J. Taylor, 2006) | Springer | DL | 5 |
| **2020** | Taylor (E. Taylor et al., 2020) | IEEE CVPRW | XAI | 5 |
| **2020** | Tjoa (Tjoa & Guan, 2020) | IEEE TNNLS | ML | 5 |
| **2020** | Uslu (Uslu et al., 2020b) | BWCCA | XAI | 5 |
| **2020** | Varshney (Varshney, 2020) | IEEE CISS | ML | 5 |
| **2018** | Volz (Volz et al., 2018) | IEEE CIG | XAI | 5 |
| **2020** | Wieringa (Wieringa, 2020) | ACM FAT | XAI | 5 |
| **2020** | Wing (Wing, 2020) | PrePrint | ML | 5 |
| **2019** | Yoon (Yoon et al., 2019) | IEEE ICCVW | XAI | 5 |
| **2019** | Zhou (Zhou & Chen, 2019) | IJCAI XAI | ML | 5 |
| **1994** | Zlatareva (N. Zlatareva & Preece, 1994) | ESA | KBS | 5 |
| **2018** | AI Now (Algorithmic Accountability Policy Tooklit, 2018) | AI Now | XAI | 4 |
| **2015** | Arifin (Arifin & Madey, 2015) | Springer | ABS | 4 |
| **2015** | Batarseh (Batarseh & Gonzalez, 2015) | AIR | KBS | 4 |
| **2007** | Brancovici (Brancovici, 2007) | IEEE CEC | XAI | 4 |
| **1987** | Castore (Castore, 1987) | NASA STI | KBS | 4 |
| **2013** | Cohen (Cohen et al., 2013) | EternalS | NLP | 4 |
| **2020** | Das (Das & Rad, 2020) | PrePrint | XAI | 4 |
| **2013** | David (David, 2013) | UCS | ABS | 4 |
| **2018** | Došilović (Došilović et al., 2018) | MIPRO | ML | 4 |
| **2000** | Edwards (Edwards, 2000) | Oxford | DS | 4 |
| **2018** | EY (Assurance in the Age of AI, 2018) | EY | ML | 4 |
| **2019** | Guidotti (Guidotti et al., 2019) | ACM CS | XAI | 4 |
| **2018** | Jilk (Jilk, 2018) | PrePrint | ABS | 4 |
| **2017** | Leibovici (Leibovici et al., 2017) | ISPRS Int J. Geo-Inf | DS | 4 |
| **2020** | Li (Li et al., 2020) | IEEE TKDE | XAI | 4 |
| **2019** | Mehrabi (Mehrabi et al., 2019) | PrePrint | ML | 4 |
| **2019** | Meskauskas (Meskauskas et al., 2020) | FUZZ-IEEE | XAI | 4 |
| **1998** | Miller (Miller, 1998) | MS | GA | 4 |
| **2019** | Nassar (Nassar et al., 2020) | WIREs DMKD | XAI | 4 |
| **1992** | Preece (Preece et al., 1992) | ESA | KBS | 4 |
| **2019** | Qiu (Qiu et al., 2019) | AS | Generic | 4 |
| **1984** | Sargent (Robert G. Sargent, 1984) | IEEE WSC | ABS | 4 |
| **2003** | Taylor (Brian J. Taylor et al., 2003) | SPIE | DL | 4 |
| **1999** | Tsai (Tsai et al., 1999) | IEEE TKDE | KBS | 4 |
| **1991** | Vinze (Vinze et al., 1991) | IM | KBS | 4 |
| **2019** | Wang (Wang et al., 2019) | ACM CHI | XAI | 4 |
| **1993** | Wells (Wells, 1993) | AAAI | KBS | 4 |
| **2018** | Zhu (J. Zhu et al., 2018) | IEEE CIG | XAI | 4 |
| **1998** | Zlatareva (N. P. Zlatareva, 1998) | DBLP | KBS | 4 |
| **2018** | Abdollahi (Abdollahi & Nasraoui, 2018) | Springer | ML | 3 |
| **1997** | Abel (Abel & Gonzalez, 1997) | FLAIRS Conference | KBS | 3 |
| **2018** | Adadi (Adadi & Berrada, 2018) | IEEE Access | XAI | 3 |
| **2018** | Agarwal (Agarwal et al., 2018) | PrePrint | Generic | 3 |
| **2016** | Amodei (Amodei et al., 2016) | PrePrint | ML | 3 |
| **2019** | Breck (Breck et al., 2019) | SysML | ML | 3 |
| **1996** | Carley (Carley, 1996) | CASOS | KBS | 3 |
| **2000** | Coenen (Coenen et al., 2000) | CUP | KBS | 3 |
| **1987** | Culbert (Culbert et al., 1987) | NASA SOAR | KBS | 3 |
| **2020** | Dağlarli (Dağlarli, 2020) | ADL | XAI | 3 |
| **1992** | Davis (Davis, 1992) | RAND | ABS | 3 |
| **2020** | Dodge (Dodge & Burnett, 2020) | ExSS-ATEC | XAI | 3 |
| **2018** | Everitt (Everitt et al., 2018) | IJCAI | AGI | 3 |
| **1991** | Gilstrap (Gilstrap, 1991) | TI | KBS | 3 |
| **2019** | Glomsrud (Glomsrud et al., 2020) | ISSAV | XAI | 3 |
| **1996** | Gonzalez (Gonzalez et al., 1996) | EAAI | KBS | 3 |
| **1997** | Harmelen (Harmelen & Teije, 1997) | EUROVAV | KBS | 3 |
| **2019** | He (Y. He et al., 2020) | PrePrint | DL | 3 |
| **2020** | Heuillet (Heuillet et al., 2020) | PrePrint | RL | 3 |
| **2009** | Hibbard (Hibbard, 2009) | AGI | AGI | 3 |
| **2019** | Israelsen (Israelsen & Ahmed, 2019) | ACM CSUR | Generic | 3 |
| **2019** | Jha (Jha et al., 2019) | NeurIPS | DL | 3 |
| **2002** | Knauf (R Knauf et al., 2002) | IEEE TSMC | KBS | 3 |
| **2017** | de Laat (de Laat, 2018) | PhilosTechnol | ML | 3 |
| **1994** | Lee (S. Lee & O'Keefe, 1994) | IEEE TSMC | KBS | 3 |
| **2004** | Liu (F. Liu & Yang, 2004) | IEEE MLC | ABS | 3 |
| **1997** | Lowry (Lowry et al., 1997) | ISMIS | Generic | 3 |
| **2012** | Martinez-Balleste (Martinez-Balleste et | IEEE SIPC | CV | 3 |
| | al., 2012) | | | |
| **2020** | Martinez-Fernandez (Martínez-Fernández et al., 2020) | PrePrint | XAI | 3 |
| **2017** | Mason (Mason et al., 2017a) | DCAART | RL | 3 |
| **1993** | Mengshoel (Mengshoel, 1993) | IEEE exp | KBS | 3 |
| **2005** | Menzies (Menzies & Pecheur, 2005) | AC | Generic | 3 |
| **2007** | Min (Feiyan Min et al., 2007) | WSC | KBS | 3 |
| **1997** | Murrell (Murrell & T. Plant, 1997) | DSS | KBS | 3 |
| **1987** | O'Keefe (O'Keefe et al., 1987) | IEEE exp | KBS | 3 |
| **2020** | Putzer (Putzer & Wozniak, 2020) | PrePrint | XAI | 3 |
| **1991** | De Raedt (De Raedt et al., 1991) | JWS | KBS | 3 |
| **2020** | Raji (Raji et al., 2020) | ACM FAT | XAI | 3 |
| **2004** | Sargent (Robert G. Sargent, 2004) | IEEE WSC | ABS | 3 |
| **1990** | Suen (Suen et al., 1990) | ESA | KBS | 3 |
| **2019** | Sun (S. C. Sun & Guo, 2020) | IEEE VTC | XAI | 3 |
| **2006** | Yilmaz (Yilmaz, 2006) | CMOT | ABS | 3 |
| **1997** | Zaidi (Zaidi & Levis, 1997) | Automatica | KBS | 3 |
| **1996** | Abel (Abel et al., 1996) | FLAIRS Conference | KBS | 2 |
| **2016** | Aitken (Aitken, 2016) | PrePrint | ABS | 2 |
| **1998** | Antoniou (Antoniou et al., 1998) | AI Magazine | KBS | 2 |
| **2019** | Arrieta (Arrieta et al., 2019) | SciDir IF | XAI | 2 |
| **2018** | Bride (Bride et al., 2018) | ICFEM | XAI | 2 |
| **2020** | Dghaym (Dghaym et al., 2020) | AU SSAV | XAI | 2 |
| **2015** | Dobson (Dobson, 2015) | JCLS | ML | 2 |
| **2018** | Hagras (Hagras, 2018) | IEEE Comp | XAI | 2 |
| **1999** | Hailu (Hailu & Sommer, 1999) | IEEE SMC | RL | 2 |
| **2020** | He (H. He et al., 2020) | IEEE IRCE | XAI | 2 |
| **2016** | Janssen (Janssen & Kuk, 2016) | GIQ | DS | 2 |
| **2020** | Kaur (Kaur et al., 2021) | NBiS | XAI | 2 |
| **2008** | Liu (F. Liu et al., 2008) | IEEE SSSC | ABS | 2 |
| **2006** | Min (Fei-yan Min et al., 2006) | ICMLC | KBS | 2 |
| **2019** | Mueller (Mueller et al., 2019) | PrePrint | XAI | 2 |
| **1996** | Nourani (Nourani, 1996) | ACM SIGSOFT | Generic | 2 |
| **2020** | Pawar (Pawar et al., 2020) | IEEE CyberSA | XAI | 2 |
| **2009** | Pèpe (Pèpe et al., 2009) | JCG | GA | 2 |
| **2013** | Pitchforth (Pitchforth, 2013) | ESA | DL | 2 |
| **2017** | Protiviti (Validation of Machine | Protiviti | ML | 2 |
| | Learning Models, 2017) | | | |
| **2010** | Sargent (Robert G. Sargent, 2010) | WSC | ABS | 2 |
| **2019** | Spada (Spada & Vincentini, 2019) | AIAI | XAI | 2 |
| **2005** | Taylor (B.J. Taylor & Darrah, 2005) | IEEE IJCNN | DL | 2 |
| **2016** | Zeigler (Zeigler & Nutaro, 2016) | JDMS | ABS | 2 |
| **2001** | Barr (Barr & Klavans, 2001) | ACL | NLP | 1 |
| **2020** | Brennen (Brennen, 2020) | ACM CHI EA | XAI | 1 |
| **2006** | Dibie-Barthélemy (Dibie-Barthélemy et al., 2006) | KBS | KBS | 1 |
| **2020** | European Commission (A European Approach to Excellence and Trust, 2020) | European Commission | XAI | 1 |
| **2000** | Gonzalez (Gonzalez & Barr, 2000) | JETAI | Generic | 1 |
| **2018** | Kaul (Kaul, 2018) | ACM AIES | ML | 1 |
| **2003** | Kurd (Kurd & Kelly, 2003) | SAFECOMP | DL | 1 |
| **2017** | Lepri (Lepri et al., 2018) | PhilosTechnology | ML | 1 |
| **2018** | Mehri (Mehri et al., 2018) | ACM ARES | DL | 1 |
| **2019** | Pocius (Pocius et al., 2019) | AAAI-19 | RL | 1 |
| **2019** | Rossi (Rossi, 2018) | JIA | XAI | 1 |
| **2010** | Schumann (Schumann et al., 2010) | NASA SCI | DL | 1 |
| **2018** | Sileno (Sileno et al., 2018) | PrePrint | XAI | 1 |
| **2019** | Varshney (Varshney, 2019) | ACM XRDS | ML | 1 |
| **2016** | Wickramage (Wickramage, 2016) | FTC | DS | 1 |

In 2018, AI papers accounted for 3% of all peer reviewed papers published worldwide (Raymond et al., 2020). The share of AI papers has grown three-fold over twenty years. Moreover, between 2010 and 2019, the total number of AI papers on arXiv increased over twenty-fold (Raymond et al., 2020). As of 2019, machine learning papers have increased most dramatically, followed by computer vision and pattern recognition. While machine learning was the most active research areas in AI, its subarea, DL have become increasing popularly in the past few years. According to GitHub, TensorFlow is the most popular free and open-source software library for AI. TensorFlow is a corporate-backed research framework, and it has been shown that, in recent years, there's noticeable trend of the emergence of such corporate-backed research frameworks. Since 2005, attendances at large AI conferences have grown significantly; NeurIPS and ICML (being the two fastest growing conferences) have over eight-fold increase. Attendances at small AI conferences have also grown over fifteen-fold starting from 2014, and the increase is highly related to the emergence of deep and reinforcement learning (Raymond et al, 2020). As the field of AI continues to grow, assurance of AI has become a more important and timely topic.

A long history of testing, validation, verification, and assurance is evident to illustrate lessons learned, pros and cons, as well as defining the future direction of AI assurance research. The next sections (4 and 5) present conclusions and recommendations for the future of AI assurance.

## 4. Recommendations and the future of AI assurance

### 4.1 The need for AI assurance

The emergence of complex, opaque, and invisible algorithms that learn from data motivated a variety of investigations, including: algorithm awareness, clarity, variance, and bias (Heuer & Breiter 2020). Algorithmic bias for instance, whether it occurs in an unintentional or intentional manner, is found to severely limit the performance of an AI model. Given AI systems provide recommendations based on data, users' faith in that the recommended outcomes are trustworthy, fair, and not biased is another critical challenge for AI assurance.

Applications of AI such as facial recognition using deep learning have become commonplace. Deep learning models are often exposed to adversarial inputs (such as deep-fakes), thus limiting their adoption and increasing their threat (Massoli et al., 2021). Unlike conventional software, aspects such as explainability (unveiling the blackbox of AI models) dictate how assurance is performed and what is needed to accomplish it. Unfortunately however, similar to the software engineering community's experience with testing, ensuring a valid and verified system is often an afterthought. Some of the classical engineering approaches would prove useful to the AI assurance community, for instance, performing testing in an incremental manner, involving users, and allocating time and budget specifically to testing, are some main lessons that ought to be considered. A worthy recent trend that might aid majorly in assurance is using AI for testing AI (i.e., deploying intelligence methods for the testing and assurance of AI methods). Additionally, from a user's perspective, recent growing questions in research that are relevant to assurance pose the following concerns: how is learning performed inside the blackbox? How is the algorithm creating its outcomes? Which dependent variables are the most influential? Is the AI algorithm dependable, safe, secure, and ethical? Besides all the previously mentioned assurance aspects, we deem the following foundational concepts as highly connected, worthy of considering by developers and AI engineers, and essential to all forms of AI assurance: (1) **Context:** refers to the scope of the system, which could be associated with a timeframe, a geographical area, specific set of users, and any other system environmental specifications (2) **Correlation:** the amount of relevance between the variables, this is usually part of exploratory analysis, however, it is key to understand which dependent variables are correlated and which ones are not, (3) **Causation:** the study of cause and effect; i.e., which variables directly cause the outcome to change (increase or decrease) in any fashion, (4) **Distribution:** whether a normal distribution is assumed or not. Data distribution of the inputted dependent variables can dictate which models are best suited for the problem at hand, and (5)

**Attribution:** aims at allocating the variables in the dataset that have the strongest influence on the outcomes of the AI algorithm.

Providing a scoring system to evaluate existing methods provides support to scholars in evaluating the field, avoiding future mistakes, and creating a system where AI scientific methods are measured and evaluated by others, a practice that is becoming increasingly rare in scientific arenas. More importantly, practitioners –in most cases– find it difficult to identify the best method for assurance relevant to their domain and subarea. We anticipate that this comprehensive review will help in that regard as well. As part of AI assurance, ethical outcomes should be evaluated, while ethical considerations might differ from one context to another, it is evident that requiring outcomes to be ethical, fair, secure, and safe necessitates the involvement of humans, and in most cases, experts from other domains. That notion qualifies AI assurance as a multidisciplinary area of investigation.

4.2 Future components of AI assurance research

In some AI subareas, there are known issues to be tackled by AI assurance, such as deep learning's sensitivity to adversarial attacks, as well as overfitting and underfitting issues in machine learning. Based on that and on the papers reviewed in this survey, it is evident that AI assurance is a necessary pursuit, but a difficult and multi-faceted area to address. However, previous experiences, successes, and failures can point us to what would work well and what is worth pursuing. Accordingly, we suggest performing and developing AI assurance by (1) domain, by (2) AI sub area, and by (3) AI goal; as a theoretical roadmap, similar to what is shown in Figure 3.

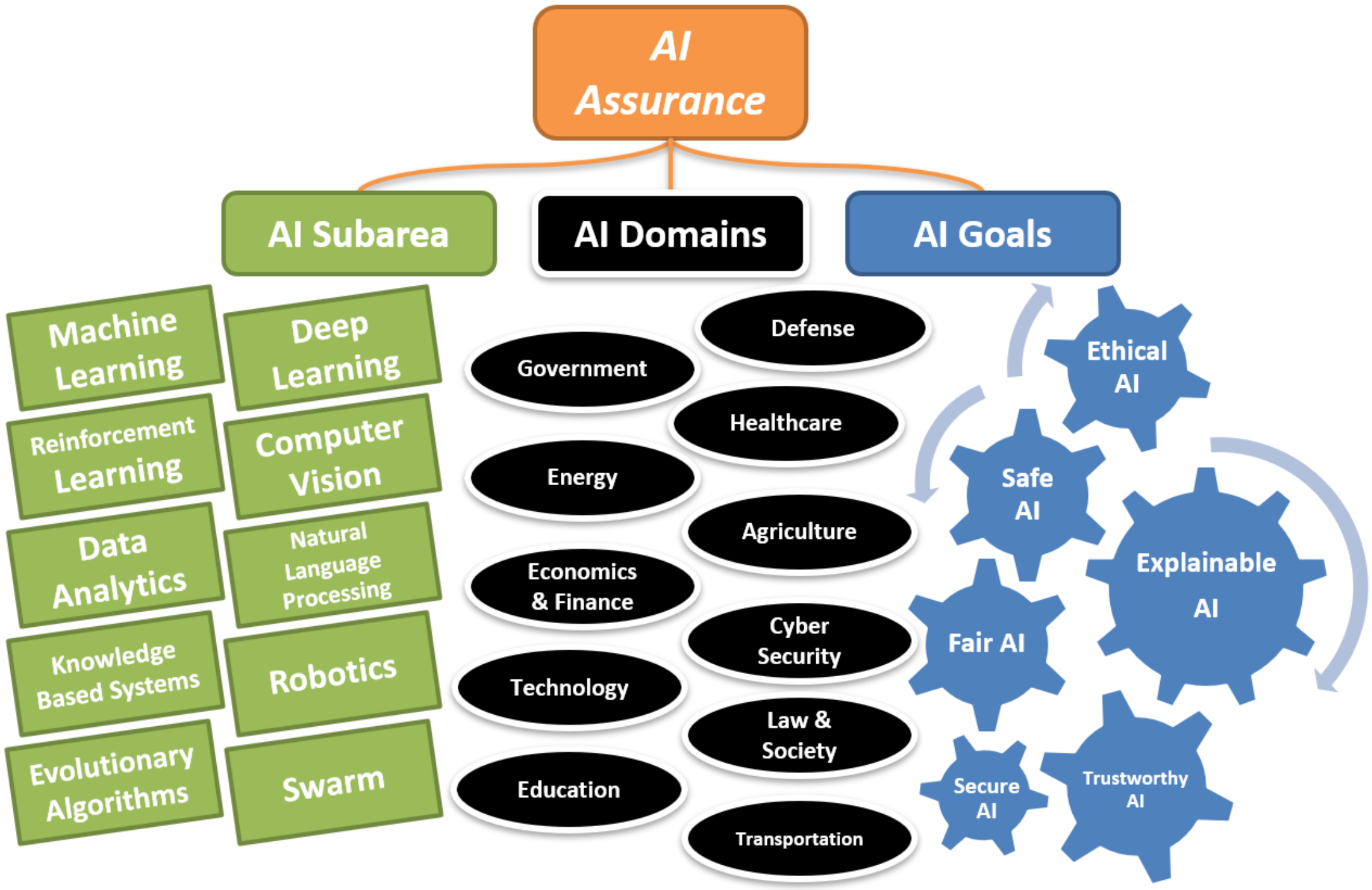

Figure 3: Three-dimensional AI assurance by subarea, domain, and goal

In some cases, such as in unsupervised learning techniques, it is difficult to know what to validate or assure (Halkidi, 2001). In such cases, the outcome is not predefined (contrary to supervised learning). Genetic algorithms and reinforcement learning have the same issue, and so in such cases, feature selection, data bias, and other data-relevant validation measures, as well as hypothesis generation and testing become more important. Additionally, different domains require different tradeoffs; trustworthiness for instance is more important when it comes to using AI in healthcare versus when its being used for revenue estimates at a private sector firm; also, AI safety is more critical in defense systems than in systems built for education or energy application.

Other surveys presented a review of AI validation and verification (Gao et al., 2016) and (Batarseh & Gonzalez, 2015), however, none was found that covered the three dimensional structure presented (by subarea, goal, and domain) like this review.

## 5. Conclusions

In AI assurance, there are other philosophical questions that are also very relevant, such as what is a valid system? What is a trustworthy outcome? When to stop testing or model learning? When to claim victory on AI safety? When to allow human intervention (and when not to)? And many other similar questions that require close attention and evaluation by the research community. The most successful methods presented in literature (scored as 8, 9, or 10), are the ones that were specific to an AI *subarea* and *goal;* additionally, ones that had done extensive theoretical and hands-on experimentation. Accordingly, we propose the following five considerations as they were evident in existing successful works when defining or applying new AI assurance methods: (1) *Data quality*: similar to assuring the outcomes, assuring the dataset and its quality mitigates issues that would eventually prevail in the AI algorithm. (2) *Specificity*: as this review concluded, the assurance methods ought to be designed to one goal and subarea of AI. (3) *Addressing invisible issues*: AI engineers should carry out assurance in a procedural manner, not as an afterthought or a process that is performed only in cases of the presence of visible issues. (4) *Automated assurance*: using manual methods for assurance would in many cases defeat the purpose. It is difficult to evaluate the validity of the assurance method itself, hence, automating the assurance process can –if done with best practices in mind– minimize error rates due to human interference. (5) *The user*: involving the user in an incremental manner is critical in expert-relevant (non-engineering) domains such as healthcare, education, economics, and other areas. Explainability is a relative and subjective matter; hence, users of the AI system can help in defining how explainability ought to be presented.

Based on all discussions presented, we assert it will be beneficial to have multi-disciplinary collaborations in the field of AI assurance. The growth of the field might need not only computer scientists and engineers to develop advanced algorithms, but also economists, physicians, biologists, lawyers, cognitive scientists, and other domain experts to unveil AI deployments to their domains, create a data-driven culture within their organizations, and ultimately enable the wide-scale adoption of assured AI systems.

---

**Declarations:**

Ethics approval and consent to participate: Not Applicable

Consent for publication: Not Applicable

Availability of data and materials: All data and materials are available under the following link: https://github.com/ferasbatarseh/AI-Assurance-Review

Competing interests: The authors declare that they have no competing interests

Funding: Not Applicable

Authors' contributions: FB designed the study, developed the visualizations, and led the effort in writing the paper; LF reviewed the paper and provided consultation on the topic; CH developed the tables, and worked on finding, arranging, and managing the papers used in the review.

Acknowledgements: Not Applicable

List of Abbreviations:
Artificial Intelligence (AI)
Third International Math and Science Study (TIMSS)
Data Science (DS)
Genetic Algorithms (GA)
Reinforcement Learning (RL)
Knowledge-Based Systems (KBS)
Computer Vision (CV)
Natural Language Processing (NLP)
Agent-Based Systems (ABS)
Machine Learning (ML)
Deep Learning (DL)
Explainable AI (XAI)

---

**Appendix 1:** All manuscripts and their detailed scores by ranking category

Columns: AI subarea: AIs; Relevance: R; Method: M; Results: Rs; Dataset: Ds; Size: Sz; Success: Sc; Limitations: L; General: G; Application: A; Comparison: C.

| Year | Author | AI.s. | R. | M. | Rs. | Ds. | Sz. | Sc. | L. | G. | A. | C. |
|---|---|---|---|---|---|---|---|---|---|---|---|---|
| 1985 | Ginsberg | KBS | 1 | 1 | 1 | 1 | 1 | 1 | 0 | 0 | 1 | 0 |
| 1987 | Castore | KBS | 1 | 1 | 0 | 0 | 0 | 0 | 1 | 0 | 1 | 0 |
| 1987 | Culbert | KBS | 1 | 0 | 0 | 0 | 0 | 0 | 1 | 0 | 1 | 0 |
| 1987 | O'Keefe | KBS | 1 | 0 | 0 | 0 | 0 | 0 | 0 | 1 | 0 | 1 |
| 1988 | Morell | KBS | 1 | 1 | 1 | 0 | 0 | 1 | 1 | 0 | 0 | 0 |
| 1988 | Sargent | ABS | 1 | 1 | 0 | 0 | 0 | 0 | 0 | 1 | 0 | 1 |
| 1989 | Becker | KBS | 1 | 1 | 1 | 0 | 0 | 1 | 0 | 0 | 1 | 0 |
| 1990 | Suen | KBS | 1 | 1 | 1 | 0 | 0 | 0 | 0 | 0 | 0 | 0 |
| 1991 | Vinze | KBS | 1 | 1 | 1 | 0 | 0 | 1 | 0 | 0 | 0 | 0 |
| 1991 | Gilstrap | KBS | 1 | 1 | 0 | 0 | 0 | 0 | 0 | 1 | 0 | 0 |
| 1991 | Raedt | KBS | 1 | 1 | 0 | 0 | 0 | 0 | 0 | 0 | 0 | 1 |
| 1992 | Andert | KBS | 1 | 1 | 1 | 0 | 0 | 1 | 0 | 0 | 0 | 1 |
| 1992 | Preece | KBS | 1 | 1 | 1 | 0 | 0 | 0 | 0 | 0 | 0 | 1 |
| 1992 | Davis | ABS | 1 | 1 | 0 | 0 | 0 | 0 | 0 | 0 | 0 | 1 |
| 1993 | Wells | KBS | 1 | 1 | 0 | 0 | 0 | 1 | 0 | 1 | 0 | 0 |
| 1993 | Mengshoel | KBS | 1 | 1 | 0 | 0 | 0 | 1 | 0 | 0 | 0 | 0 |
| 1994 | Zlatareva | KBS | 1 | 1 | 1 | 0 | 0 | 1 | 0 | 0 | 1 | 0 |
| 1994 | Lee | KBS | 1 | 1 | 0 | 0 | 0 | 0 | 0 | 0 | 0 | 1 |
| 1995 | Lockwood | KBS | 1 | 1 | 1 | 0 | 0 | 1 | 0 | 0 | 1 | 0 |
| 1995 | Singer | DS | 1 | 1 | 1 | 1 | 0 | 1 | 0 | 0 | 0 | 0 |
| 1996 | Martin | DS | 0 | 1 | 1 | 0 | 0 | 1 | 0 | 1 | 1 | 1 |
| 1996 | Carley | KBS | 1 | 0 | 0 | 0 | 0 | 0 | 0 | 1 | 0 | 1 |
| 1996 | Gonzalez | KBS | 1 | 1 | 0 | 0 | 0 | 1 | 0 | 0 | 0 | 0 |
| 1996 | Abel | KBS | 1 | 1 | 0 | 0 | 0 | 0 | 0 | 0 | 0 | 0 |
| 1996 | Nourani | Generic | 1 | 1 | 0 | 0 | 0 | 0 | 0 | 0 | 0 | 0 |
| 1997 | Jones | GA | 0 | 1 | 1 | 0 | 0 | 1 | 0 | 0 | 1 | 1 |
| 1997 | Abel | KBS | 1 | 1 | 1 | 0 | 0 | 0 | 0 | 0 | 0 | 0 |
| 1997 | Harmelen | KBS | 1 | 1 | 0 | 0 | 0 | 0 | 0 | 0 | 0 | 1 |
| 1997 | Lowry | Generic | 1 | 1 | 0 | 0 | 0 | 0 | 0 | 0 | 1 | 0 |
| 1997 | Murrell | KBS | 1 | 0 | 0 | 0 | 0 | 0 | 0 | 1 | 0 | 1 |
| 1997 | Zaidi | KBS | 1 | 1 | 0 | 0 | 0 | 1 | 0 | 0 | 0 | 0 |
| 1998 | Miller | GA | 1 | 1 | 1 | 0 | 0 | 1 | 0 | 0 | 0 | 0 |
| 1998 | Zlatareva | KBS | 1 | 1 | 1 | 0 | 0 | 1 | 0 | 0 | 0 | 0 |
| 1998 | Antoniou | KBS | 0 | 1 | 0 | 0 | 0 | 0 | 0 | 0 | 0 | 1 |
| 1999 | Thomas | GA | 1 | 1 | 1 | 1 | 1 | 1 | 0 | 0 | 0 | 1 |
| 1999 | Tsai | KBS | 1 | 1 | 0 | 0 | 0 | 0 | 0 | 0 | 1 | 1 |

| | | | | | | | | | | | |
|---|---|---|---|---|---|---|---|---|---|---|---|
| 1999 | Hailu | RL | 0 | 1 | 1 | 0 | 0 | 0 | 0 | 0 | 0 | 0 |
| 2000 | Mosqueira-Rey | KBS | 1 | 1 | 1 | 0 | 0 | 1 | 0 | 1 | 1 | 0 |
| 2000 | Marcos | KBS | 1 | 0 | 1 | 0 | 0 | 1 | 0 | 1 | 1 | 0 |
| 2000 | Onoyama | KBS | 1 | 1 | 1 | 0 | 0 | 1 | 0 | 0 | 1 | 0 |
| 2000 | Edwards | DS | 1 | 0 | 0 | 0 | 0 | 0 | 0 | 1 | 1 | 1 |
| 2000 | Coenen | KBS | 0 | 0 | 0 | 0 | 0 | 0 | 0 | 1 | 1 | 1 |
| 2000 | Gonzalez | Generic | 0 | 0 | 0 | 0 | 0 | 0 | 0 | 1 | 0 | 0 |
| 2001 | Berndt | DS | 1 | 1 | 1 | 1 | 1 | 1 | 0 | 0 | 1 | 0 |
| 2001 | Halkidi | ML | 1 | 1 | 1 | 0 | 0 | 1 | 0 | 0 | 1 | 1 |
| 2001 | Barr | NLP | 0 | 0 | 0 | 0 | 0 | 0 | 0 | 0 | 1 | 0 |
| 2002 | Wu | KBS | 1 | 1 | 0 | 0 | 0 | 1 | 1 | 0 | 1 | 1 |
| 2002 | Knauf | KBS | 1 | 1 | 0 | 0 | 0 | 1 | 0 | 0 | 0 | 0 |
| 2003 | Schumann | DL | 1 | 1 | 1 | 0 | 0 | 0 | 0 | 0 | 1 | 1 |
| 2003 | Taylor | DL | 1 | 1 | 0 | 0 | 0 | 0 | 0 | 1 | 0 | 1 |
| 2003 | Kurd | DL | 0 | 0 | 0 | 0 | 0 | 0 | 0 | 1 | 0 | 0 |
| 2004 | Liu | ABS | 1 | 0 | 0 | 0 | 0 | 0 | 0 | 1 | 0 | 1 |
| 2004 | Sargent | ABS | 1 | 0 | 0 | 0 | 0 | 0 | 0 | 1 | 0 | 1 |
| 2005 | Liu | DL | 1 | 1 | 1 | 1 | 0 | 1 | 0 | 0 | 0 | 1 |
| 2005 | Menzies | Generic | 1 | 1 | 0 | 0 | 0 | 0 | 0 | 1 | 0 | 0 |
| 2005 | Taylor | DL | 1 | 1 | 0 | 0 | 0 | 0 | 0 | 0 | 0 | 0 |
| 2006 | Forster | AGI | 1 | 1 | 1 | 1 | 1 | 1 | 0 | 0 | 0 | 1 |
| 2006 | Taylor | DL | 1 | 0 | 1 | 0 | 0 | 0 | 0 | 1 | 1 | 1 |
| 2006 | Yilmaz | ABS | 1 | 1 | 0 | 0 | 0 | 0 | 0 | 0 | 0 | 1 |
| 2006 | Min | KBS | 1 | 1 | 0 | 0 | 0 | 0 | 0 | 0 | 0 | 0 |
| 2006 | Dibie-Barthélemy | KBS | 0 | 0 | 0 | 0 | 0 | 0 | 0 | 0 | 0 | 1 |
| 2007 | Martín-Guerrero | RL | 1 | 1 | 1 | 1 | 0 | 1 | 0 | 0 | 1 | 0 |
| 2007 | Knauf | KBS | 1 | 1 | 1 | 1 | 0 | 0 | 0 | 0 | 0 | 1 |
| 2007 | Brancovici | XAI | 1 | 1 | 0 | 0 | 0 | 0 | 0 | 0 | 1 | 1 |
| 2007 | Min | KBS | 1 | 1 | 0 | 0 | 0 | 0 | 0 | 0 | 1 | 0 |
| 2008 | Peppler | DS | 1 | 1 | 1 | 1 | 1 | 1 | 1 | 0 | 1 | 0 |
| 2008 | Liu | ABS | 1 | 1 | 0 | 0 | 0 | 0 | 0 | 0 | 0 | 0 |
| 2009 | Tadj | KBS | 1 | 1 | 1 | 1 | 0 | 1 | 1 | 0 | 0 | 1 |
| 2009 | Hibbard | AGI | 0 | 1 | 1 | 0 | 0 | 1 | 0 | 0 | 0 | 0 |
| 2009 | Pèpe | GA | 0 | 1 | 1 | 0 | 0 | 0 | 0 | 0 | 0 | 0 |
| 2010 | Bone | RL | 1 | 1 | 1 | 1 | 1 | 1 | 0 | 0 | 1 | 0 |
| 2010 | Winkler | CV | 1 | 1 | 1 | 0 | 0 | 1 | 0 | 0 | 1 | 1 |
| 2010 | Dong | GA | 1 | 1 | 1 | 0 | 0 | 1 | 0 | 0 | 0 | 1 |
| 2010 | Niazi | ABS | 1 | 1 | 1 | 1 | 0 | 1 | 0 | 0 | 0 | 0 |
| 2010 | Sargent | ABS | 0 | 0 | 0 | 0 | 0 | 0 | 0 | 1 | 0 | 1 |
| 2010 | Schumann | DL | 0 | 0 | 0 | 0 | 0 | 0 | 0 | 0 | 0 | 1 |
| 2012 | Cohen | NLP | 0 | 1 | 1 | 1 | 0 | 1 | 0 | 0 | 0 | 0 |
| 2012 | Martinez- | CV | 1 | 0 | 0 | 0 | 0 | 0 | 0 | 0 | 1 | 1 |
| | Balleste | | | | | | | | | | | |
| 2013 | Batarseh | KBS | 1 | 1 | 1 | 1 | 1 | 1 | 0 | 0 | 1 | 0 |
| 2013 | Sargent | ABS | 1 | 1 | 1 | 0 | 0 | 0 | 0 | 0 | 1 | 1 |
| 2013 | David | ABS | 1 | 1 | 0 | 0 | 0 | 0 | 0 | 1 | 0 | 1 |
| 2013 | Pitchforth | DL | 1 | 1 | 0 | 0 | 0 | 0 | 0 | 0 | 0 | 0 |
| 2014 | Ali | DS | 1 | 1 | 1 | 1 | 1 | 1 | 1 | 0 | 0 | 1 |
| 2015 | He | GA | 1 | 1 | 1 | 0 | 0 | 1 | 1 | 0 | 1 | 1 |
| 2015 | Rovcanin | RL | 1 | 1 | 1 | 1 | 0 | 1 | 0 | 0 | 1 | 1 |
| 2015 | Goodfellow | ML | 1 | 1 | 1 | 0 | 0 | 1 | 0 | 0 | 0 | 1 |
| 2015 | Arifin | ABS | 1 | 0 | 0 | 0 | 0 | 0 | 0 | 1 | 1 | 1 |
| 2015 | Batarseh | KBS | 1 | 0 | 0 | 0 | 0 | 0 | 0 | 1 | 1 | 1 |
| 2015 | Dobson | ML | 1 | 0 | 0 | 0 | 0 | 0 | 0 | 0 | 0 | 1 |
| 2016 | Veeramachaneni | DS | 1 | 1 | 1 | 1 | 1 | 1 | 1 | 1 | 1 | 0 |
| 2016 | Gao | DS | 1 | 0 | 1 | 1 | 1 | 1 | 0 | 1 | 1 | 1 |
| 2016 | Gulshan | CV | 1 | 1 | 1 | 1 | 1 | 1 | 0 | 0 | 1 | 1 |
| 2016 | Heaney | GA | 1 | 1 | 1 | 1 | 1 | 1 | 1 | 0 | 1 | 0 |
| 2016 | Aitken | ABS | 1 | 1 | 1 | 1 | 1 | 1 | 0 | 0 | 0 | 1 |
| 2016 | Celis | ML | 1 | 1 | 1 | 1 | 1 | 1 | 0 | 0 | 1 | 0 |
| 2016 | Jiang | RL | 0 | 1 | 1 | 1 | 1 | 1 | 0 | 0 | 1 | 1 |
| 2016 | Kianifar | GA | 1 | 1 | 1 | 1 | 1 | 1 | 0 | 0 | 1 | 0 |
| 2016 | Jilk | ABS | 1 | 0 | 1 | 0 | 0 | 1 | 0 | 1 | 0 | 0 |
| 2016 | Amodei | ML | 1 | 0 | 0 | 0 | 0 | 0 | 0 | 1 | 0 | 1 |
| 2016 | Aitken | ABS | 1 | 1 | 0 | 0 | 0 | 0 | 0 | 0 | 0 | 0 |
| 2016 | Janssen | DS | 0 | 0 | 0 | 0 | 0 | 0 | 0 | 1 | 0 | 1 |
| 2016 | Zeigler | ABS | 1 | 1 | 0 | 0 | 0 | 0 | 0 | 0 | 0 | 0 |
| 2016 | Wickramage | DS | 1 | 0 | 0 | 0 | 0 | 0 | 0 | 0 | 0 | 0 |
| 2017 | Liang | DS | 1 | 1 | 1 | 1 | 0 | 1 | 0 | 0 | 1 | 1 |
| 2017 | Xu | DL | 1 | 1 | 1 | 1 | 1 | 1 | 0 | 0 | 0 | 1 |
| 2017 | Mason | RL | 1 | 1 | 1 | 0 | 0 | 1 | 0 | 0 | 0 | 1 |
| 2017 | Leibovici | DS | 1 | 1 | 0 | 0 | 0 | 0 | 0 | 1 | 0 | 1 |
| 2017 | Laat | ML | 0 | 1 | 0 | 0 | 0 | 0 | 0 | 1 | 0 | 1 |
| 2017 | Mason | RL | 1 | 1 | 0 | 0 | 0 | 0 | 0 | 0 | 0 | 1 |
| 2017 | Lepri | ML | 0 | 0 | 0 | 0 | 0 | 0 | 0 | 0 | 0 | 1 |
| 2018 | Wei | RL | 1 | 1 | 1 | 1 | 1 | 1 | 1 | 0 | 1 | 1 |
| 2018 | Alves | ABS | 1 | 1 | 1 | 1 | 1 | 1 | 0 | 0 | 1 | 1 |
| 2018 | Elsayed | CV | 1 | 1 | 1 | 1 | 1 | 1 | 0 | 0 | 0 | 1 |
| 2018 | Go | DL | 1 | 1 | 1 | 1 | 1 | 1 | 0 | 0 | 1 | 0 |
| 2018 | Mason | RL | 1 | 1 | 1 | 0 | 0 | 1 | 1 | 1 | 0 | 1 |
| 2018 | Murray | XAI | 1 | 1 | 1 | 1 | 1 | 1 | 0 | 0 | 0 | 1 |
| 2018 | Pynadath | ML | 1 | 1 | 1 | 0 | 0 | 1 | 1 | 0 | 1 | 1 |
| 2018 | Stock | CV | 1 | 1 | 1 | 1 | 1 | 1 | 0 | 0 | 1 | 0 |
| 2018 | Cao | ML | 1 | 1 | 1 | 1 | 0 | 1 | 0 | 0 | 0 | 1 |
| 2018 | Ruan | DL | 1 | 0 | 1 | 1 | 0 | 1 | 1 | 0 | 0 | 1 |
| 2018 | Antunes | ML | 1 | 1 | 1 | 0 | 0 | 1 | 1 | 0 | 0 | 0 |
| 2018 | Volz | XAI | 0 | 1 | 1 | 0 | 0 | 1 | 0 | 0 | 1 | 1 |
| 2018 | AI Now | XAI | 1 | 1 | 0 | 0 | 0 | 0 | 0 | 1 | 1 | 0 |
| 2018 | Došilović | ML | 1 | 0 | 0 | 0 | 0 | 0 | 0 | 1 | 1 | 1 |
| 2018 | EY | ML | 1 | 0 | 0 | 0 | 0 | 0 | 0 | 1 | 1 | 1 |
| 2018 | Guidotti | XAI | 1 | 0 | 0 | 0 | 0 | 0 | 0 | 1 | 1 | 1 |
| 2018 | Zhu | XAI | 1 | 0 | 0 | 0 | 0 | 0 | 0 | 1 | 1 | 1 |
| 2018 | Abdollahi | ML | 1 | 0 | 0 | 0 | 0 | 0 | 0 | 1 | 0 | 1 |
| 2018 | Adadi | XAI | 1 | 0 | 0 | 0 | 0 | 0 | 0 | 1 | 0 | 1 |
| 2018 | Agarwal | Generic | 1 | 1 | 0 | 0 | 0 | 0 | 0 | 0 | 0 | 1 |
| 2018 | Everitt | AGI | 1 | 0 | 0 | 0 | 0 | 0 | 0 | 1 | 0 | 1 |
| 2018 | Bride | XAI | 1 | 1 | 0 | 0 | 0 | 0 | 0 | 0 | 0 | 0 |
| 2018 | Hagras | XAI | 1 | 0 | 0 | 0 | 0 | 0 | 0 | 0 | 0 | 1 |
| 2018 | Kaul | ML | 1 | 0 | 0 | 0 | 0 | 0 | 0 | 0 | 0 | 0 |
| 2018 | Mehri | DL | 0 | 0 | 0 | 0 | 0 | 0 | 0 | 0 | 0 | 1 |
| 2018 | Sileno | XAI | 1 | 0 | 0 | 0 | 0 | 0 | 0 | 0 | 0 | 0 |
| 2019 | Tao | Generic | 1 | 1 | 1 | 1 | 1 | 1 | 1 | 1 | 1 | 1 |
| 2019 | Kaur | XAI | 1 | 1 | 1 | 1 | 1 | 1 | 1 | 0 | 1 | 1 |
| 2019 | Batarseh | DS | 1 | 1 | 1 | 1 | 1 | 1 | 0 | 0 | 1 | 1 |
| 2019 | Huber | RL | 1 | 1 | 1 | 1 | 1 | 1 | 1 | 0 | 0 | 1 |
| 2019 | Keneni | XAI | 1 | 1 | 1 | 1 | 1 | 1 | 1 | 0 | 1 | 0 |
| 2019 | Maloca | DL | 1 | 1 | 1 | 1 | 1 | 1 | 0 | 0 | 1 | 1 |
| 2019 | Barredo-Arrieta | XAI | 1 | 1 | 1 | 1 | 1 | 0 | 0 | 1 | 1 | 0 |
| 2019 | Chittajallu | XAI | 1 | 1 | 1 | 1 | 0 | 1 | 1 | 0 | 0 | 1 |
| 2019 | Ferreyra | XAI | 1 | 1 | 1 | 0 | 0 | 1 | 0 | 1 | 1 | 1 |
| 2019 | Lee | XAI | 1 | 1 | 1 | 1 | 1 | 1 | 0 | 0 | 0 | 1 |
| 2019 | Naqa | ML | 1 | 1 | 1 | 1 | 0 | 1 | 0 | 0 | 1 | 1 |
| 2019 | Prentzas | XAI | 1 | 1 | 1 | 1 | 1 | 1 | 0 | 0 | 1 | 0 |
| 2019 | Bellamy | XAI | 1 | 1 | 1 | 0 | 0 | 1 | 1 | 1 | 0 | 0 |
| 2019 | Beyret | RL | 1 | 1 | 1 | 0 | 0 | 1 | 1 | 0 | 1 | 0 |
| 2019 | Madumal | RL | 1 | 1 | 1 | 0 | 0 | 1 | 0 | 0 | 1 | 1 |
| 2019 | Schlegel | XAI | 1 | 1 | 1 | 1 | 1 | 1 | 0 | 0 | 0 | 0 |
| 2019 | Vabalas | ML | 1 | 1 | 1 | 1 | 0 | 1 | 0 | 0 | 1 | 0 |
| 2019 | Zhu | RL | 1 | 1 | 1 | 0 | 0 | 1 | 1 | 0 | 0 | 1 |
| 2019 | Chen | RL | 1 | 1 | 0 | 0 | 0 | 0 | 0 | 1 | 1 | 1 |
| 2019 | Cruz | RL | 1 | 1 | 1 | 0 | 0 | 1 | 0 | 0 | 0 | 1 |
| 2019 | Dupuis | XAI | 1 | 1 | 1 | 0 | 0 | 1 | 0 | 0 | 0 | 1 |
| 2019 | Joo | RL | 1 | 1 | 1 | 0 | 0 | 1 | 1 | 0 | 0 | 0 |
| 2019 | Ren | DL | 0 | 1 | 1 | 0 | 0 | 1 | 1 | 0 | 1 | 0 |
| 2019 | Srivastava | NLP | 1 | 1 | 1 | 0 | 0 | 1 | 0 | 0 | 0 | 1 |
| 2019 | Uslu | XAI | 1 | 1 | 1 | 0 | 0 | 1 | 0 | 0 | 0 | 1 |
| 2019 | Yoon | XAI | 1 | 1 | 1 | 0 | 0 | 1 | 0 | 0 | 0 | 1 |
| 2019 | Zhou | ML | 1 | 1 | 1 | 0 | 0 | 1 | 0 | 0 | 0 | 1 |
| 2019 | Mehrabi | ML | 1 | 0 | 0 | 0 | 0 | 0 | 0 | 1 | 1 | 1 |
| 2019 | Meskauskas | XAI | 1 | 1 | 1 | 0 | 0 | 1 | 0 | 0 | 0 | 0 |
| 2019 | Nassar | XAI | 1 | 1 | 0 | 0 | 0 | 0 | 0 | 0 | 1 | 1 |
| 2019 | Qiu | Generic | 1 | 0 | 0 | 0 | 0 | 0 | 0 | 1 | 1 | 1 |
| 2019 | Wang | XAI | 1 | 1 | 0 | 0 | 0 | 0 | 0 | 1 | 0 | 1 |
| 2019 | Breck | ML | 1 | 1 | 0 | 0 | 0 | 0 | 0 | 0 | 1 | 0 |
| 2019 | Glomsrud | XAI | 1 | 0 | 0 | 0 | 0 | 0 | 0 | 0 | 1 | 1 |
| 2019 | He | DL | 1 | 0 | 0 | 0 | 0 | 0 | 0 | 1 | 0 | 1 |
| 2019 | Israelsen | Generic | 1 | 0 | 0 | 0 | 0 | 0 | 0 | 1 | 0 | 1 |
| 2019 | Jha | DL | 1 | 0 | 1 | 0 | 0 | 0 | 0 | 0 | 0 | 1 |
| 2019 | Sun | XAI | 0 | 1 | 1 | 0 | 0 | 1 | 0 | 0 | 0 | 0 |
| 2019 | Dghaym | XAI | 0 | 0 | 0 | 0 | 0 | 0 | 0 | 0 | 1 | 1 |
| 2019 | Mueller | XAI | 1 | 0 | 0 | 0 | 0 | 0 | 0 | 0 | 0 | 1 |
| 2019 | Protiviti | ML | 0 | 0 | 0 | 0 | 0 | 0 | 0 | 1 | 0 | 1 |
| 2019 | Spada | XAI | 0 | 1 | 0 | 0 | 0 | 0 | 0 | 0 | 1 | 0 |
| 2019 | Pocius | RL | 1 | 0 | 0 | 0 | 0 | 0 | 0 | 0 | 0 | 0 |
| 2019 | Rossi | XAI | 1 | 0 | 0 | 0 | 0 | 0 | 0 | 0 | 0 | 0 |
| 2019 | Varshney | ML | 0 | 0 | 0 | 0 | 0 | 0 | 0 | 1 | 0 | 0 |
| 2020 | D'Alterio | XAI | 1 | 1 | 1 | 1 | 1 | 1 | 1 | 1 | 1 | 1 |
| 2020 | Anderson | RL | 1 | 1 | 1 | 1 | 1 | 1 | 0 | 1 | 1 | 1 |
| 2020 | Birkenbihl | ML | 1 | 1 | 1 | 1 | 1 | 1 | 1 | 0 | 1 | 1 |
| 2020 | Checco | DS | 1 | 1 | 1 | 1 | 1 | 1 | 0 | 1 | 1 | 1 |
| 2020 | Chen | XAI | 1 | 1 | 1 | 1 | 1 | 1 | 1 | 0 | 1 | 1 |
| 2020 | EASA | DL | 1 | 1 | 1 | 1 | 1 | 1 | 0 | 1 | 1 | 1 |
| 2020 | Kulkarni | DS | 1 | 1 | 1 | 1 | 1 | 1 | 0 | 1 | 1 | 1 |
| 2020 | Kuppa | XAI | 1 | 1 | 1 | 1 | 1 | 1 | 1 | 0 | 1 | 1 |
| 2020 | Kuzlu | XAI | 1 | 1 | 1 | 1 | 1 | 1 | 0 | 1 | 1 | 1 |
| 2020 | Spinner | XAI | 1 | 1 | 1 | 1 | 1 | 1 | 0 | 1 | 1 | 1 |
| 2020 | Winkel | RL | 1 | 1 | 1 | 1 | 1 | 1 | 1 | 0 | 1 | 1 |
| 2020 | Gardiner | ML | 1 | 1 | 1 | 1 | 1 | 1 | 0 | 0 | 1 | 1 |
| 2020 | Guo | XAI | 1 | 1 | 1 | 1 | 1 | 1 | 1 | 0 | 1 | 0 |
| 2020 | Han | XAI | 1 | 1 | 1 | 1 | 1 | 1 | 0 | 0 | 1 | 1 |
| 2020 | Kohlbrenner | XAI | 1 | 1 | 1 | 1 | 1 | 1 | 1 | 0 | 0 | 1 |
| 2020 | Malolan | XAI | 1 | 1 | 1 | 1 | 1 | 1 | 1 | 0 | 1 | 0 |
| 2020 | Payrovnaziri | ML | 1 | 1 | 0 | 1 | 1 | 0 | 1 | 1 | 1 | 1 |
| 2020 | Sequeira | RL | 1 | 1 | 1 | 1 | 1 | 1 | 0 | 0 | 1 | 1 |
| 2020 | Sivamani | DL | 1 | 1 | 1 | 1 | 1 | 1 | 0 | 0 | 1 | 1 |
| 2020 | Tan | XAI | 1 | 1 | 1 | 1 | 1 | 1 | 0 | 0 | 1 | 1 |
| 2020 | Tao | XAI | 1 | 1 | 1 | 1 | 1 | 1 | 1 | 0 | 1 | 0 |
| 2020 | Welch | DL | 1 | 1 | 1 | 1 | 1 | 1 | 0 | 0 | 1 | 1 |
| 2020 | Xiao | DL | 1 | 1 | 1 | 1 | 1 | 1 | 0 | 0 | 1 | 1 |
| 2020 | Halliwell | DL | 1 | 1 | 1 | 1 | 1 | 1 | 0 | 0 | 0 | 1 |
| 2020 | Heuer | ML | 1 | 1 | 1 | 1 | 1 | 1 | 0 | 0 | 0 | 1 |

| | | | | | | | | | | | |
|---|---|---|---|---|---|---|---|---|---|---|---|
| 2020 | Kaur | XAI | 1 | 1 | 1 | 1 | 1 | 1 | 0 | 0 | 1 | 0 |
| 2020 | Mackowiak | CV | 1 | 1 | 1 | 1 | 0 | 1 | 0 | 0 | 1 | 1 |
| 2020 | Ragot | ML | 1 | 1 | 1 | 1 | 1 | 1 | 0 | 0 | 1 | 0 |
| 2020 | Rotman | RL | 1 | 1 | 1 | 1 | 1 | 0 | 0 | 0 | 1 | 1 |
| 2020 | Sarathy | XAI | 1 | 1 | 1 | 1 | 0 | 1 | 0 | 0 | 1 | 1 |
| 2020 | Uslu | XAI | 0 | 1 | 1 | 1 | 1 | 1 | 0 | 0 | 1 | 1 |
| 2020 | Cruz | RL | 1 | 1 | 1 | 0 | 0 | 1 | 1 | 0 | 0 | 1 |
| 2020 | He | RL | 1 | 1 | 1 | 1 | 1 | 0 | 0 | 0 | 1 | 0 |
| 2020 | Islam | XAI | 0 | 1 | 1 | 1 | 1 | 0 | 0 | 0 | 1 | 1 |
| 2020 | Mynuddin | RL | 1 | 1 | 1 | 1 | 0 | 1 | 0 | 0 | 1 | 0 |
| 2020 | Puiutta | RL | 1 | 1 | 0 | 0 | 0 | 0 | 1 | 1 | 1 | 1 |
| 2020 | Toreini | ML | 1 | 0 | 0 | 0 | 1 | 1 | 1 | 1 | 0 | 1 |
| 2020 | Toreini | ML | 1 | 0 | 0 | 0 | 1 | 1 | 1 | 1 | 0 | 1 |
| 2020 | Diallo | XAI | 1 | 1 | 1 | 1 | 0 | 1 | 0 | 0 | 0 | 0 |
| 2020 | Guo | XAI | 1 | 0 | 0 | 0 | 0 | 1 | 0 | 1 | 1 | 1 |
| 2020 | Haverinen | XAI | 0 | 1 | 1 | 1 | 0 | 1 | 0 | 0 | 0 | 1 |
| 2020 | Katell | XAI | 1 | 1 | 0 | 0 | 0 | 0 | 1 | 0 | 1 | 1 |
| 2020 | Murray | XAI | 0 | 1 | 1 | 1 | 1 | 0 | 0 | 0 | 1 | 0 |
| 2020 | Taylor | XAI | 1 | 1 | 1 | 1 | 1 | 0 | 0 | 0 | 0 | 0 |
| 2020 | Tjoa | ML | 1 | 0 | 0 | 0 | 1 | 0 | 1 | 1 | 0 | 1 |
| 2020 | Varshney | ML | 1 | 1 | 0 | 0 | 0 | 0 | 1 | 1 | 0 | 1 |
| 2020 | Wieringa | XAI | 1 | 0 | 0 | 0 | 1 | 0 | 1 | 1 | 0 | 1 |
| 2020 | Wing | ML | 1 | 0 | 0 | 0 | 1 | 0 | 1 | 1 | 0 | 1 |
| 2020 | Das | XAI | 1 | 0 | 0 | 0 | 0 | 0 | 1 | 1 | 0 | 1 |
| 2020 | Li | XAI | 0 | 0 | 0 | 0 | 1 | 0 | 1 | 1 | 0 | 1 |
| 2020 | Dağlarli | XAI | 1 | 0 | 0 | 0 | 0 | 0 | 0 | 1 | 0 | 1 |
| 2020 | Dodge | XAI | 1 | 1 | 0 | 0 | 0 | 0 | 0 | 0 | 0 | 1 |
| 2020 | Heuillet | RL | 1 | 0 | 0 | 0 | 0 | 0 | 0 | 1 | 0 | 1 |
| 2020 | Martinez-Fernandez | XAI | 1 | 1 | 0 | 0 | 0 | 0 | 0 | 0 | 1 | 0 |
| 2020 | Putzer | XAI | 0 | 1 | 0 | 0 | 0 | 0 | 0 | 0 | 1 | 1 |
| 2020 | Raji | XAI | 1 | 0 | 0 | 0 | 0 | 0 | 0 | 1 | 1 | 0 |
| 2020 | Arrieta | XAI | 1 | 0 | 0 | 0 | 0 | 0 | 0 | 0 | 0 | 1 |
| 2020 | He | XAI | 1 | 0 | 0 | 0 | 0 | 0 | 0 | 1 | 0 | 0 |
| 2020 | Kaur | XAI | 1 | 0 | 0 | 0 | 0 | 0 | 0 | 0 | 0 | 1 |
| 2020 | Pawar | XAI | 0 | 1 | 0 | 0 | 0 | 0 | 0 | 0 | 0 | 1 |
| 2020 | Brennen | XAI | 0 | 0 | 0 | 0 | 0 | 0 | 0 | 1 | 0 | 0 |
| 2020 | European Commission | XAI | 0 | 0 | 0 | 0 | 0 | 0 | 0 | 1 | 0 | 0 |
| 2021 | Massoli | DL | 1 | 1 | 1 | 1 | 1 | 1 | 1 | 0 | 1 | 1 |